\documentclass[runningheads]{llncs}
\usepackage[T1]{fontenc}
\usepackage{graphicx}
\usepackage{amsmath,amsfonts,bm}
\usepackage{booktabs}
\usepackage{subcaption}
\begin{document}
\title{Hypergame Rationalisability: Solving Agent Misalignment In Strategic Play}
%
%
\author{Vince Trencsenyi\inst{1}\orcidID{0009-0009-4560-7571}}
%
\authorrunning{V. Trencsenyi}
%
\institute{Royal Holloway University of London, Egham TW20 0EX, United Kingdom \\
\email{vince.trencsenyi@rhul.ac.uk}}
%


\maketitle              
\begin{abstract}
Differences in perception, information asymmetries, and\\ bounded rationality lead game-theoretic players to derive a private,\\ subjective view of the game that may diverge from the underlying\\ ``ground‑truth'' scenario and may be misaligned with other players' interpretations. While typical game-theoretic assumptions often overlook such heterogeneity, hypergame theory provides the mathematical framework to reason about mismatched mental models. Although hypergames have recently gained traction in dynamic applications concerning uncertainty, their practical adoption in multi-agent system research has been hindered by the lack of a unifying, formal, and practical representation language, as well as scalable algorithms for managing complex hypergame structures and equilibria. Our work addresses this gap by introducing a declarative, logic-based domain-specific language for encoding hypergame structures and hypergame solution concepts. Leveraging answer-set programming, we develop an automated pipeline for instantiating hypergame structures and running our novel hypergame rationalisation procedure, a mechanism for finding belief structures that justify seemingly irrational outcomes. The proposed language establishes a unifying formalism for hypergames and serves as a foundation for developing nuanced, belief-based heterogeneous reasoners, offering a verifiable context with logical guarantees. Together, these contributions establish the connection between hypergame theory, multi-agent systems, and strategic AI. 

\end{abstract}
\section{Introduction}

Agents -- human or artificial -- rarely share a perfectly aligned understanding of the strategic situations they inhabit. While game theory provides a mathematical framework for modelling and analysing decision-making in such strategic situations, typical game-theoretic models rely on simplifying assumptions about players' rationality and interpretation of payoffs, omitting factors of heterogeneity~\cite{Rasmusen1990-RASGAI}. While these abstractions serve the tractability of game-theoretic models, they have also been subject to critique, stating that the extensive generalisation omits the social context of the interaction, dehumanising the models and thus creating a gap between theory and practical applications~\cite{Morgenstern1964,Rapoport1962}. Driven by similar motivations, Bennett~\cite{Bennett1977} proposed hypergame theory as an extension of game-theoretic concepts, relaxing the assumption of player homogeneity and allowing players to develop subjective games that capture their individual interpretations of the interaction based on their perceptions and beliefs.

Agent-based applications provide a natural platform for scenarios that encompass competitive or cooperative interactions within societies of actors~\cite{wooldridge2009introduction}. Given that game theory offers a set of analytical tools for the same class of scenarios, Multi-Agent Systems (MAS) and game theory go hand-in-hand~\cite{michel2018multi,parsons2002game,PENDHARKAR2012273,tradingagentcompetition}: MAS benefit from adopting game-theoretic interaction models, while game theory can rely on MAS for developing social simulations where strategic behaviour can be evaluated systematically. In this context, as agent misalignment remains a top-priority challenge to be solved~\cite{kierans2024quantifyingmisalignmentagentssociotechnical,Mechergui2024GoalAR,pearson2022aibook,sarkadi2019towards}, developing realistic game-theoretic models as an interaction protocol for multi-agent based simulations has been receiving increasing research focus~\cite{Sasaki2021multiagaentdecisionsystems}. Consequently, while hypergame theory was initially developed primarily as an analytical framework, there has been a recent increasing tendency of hypergame-based multi-agent applications, most frequently targeting models of misaligned perceptions and deception~\cite{trencsenyi2025hypergames}.

In this work, we focus on MAS applications that directly integrate hypergames into their interaction mechanisms. In particular,~\cite{trencsenyi2025approximating,trencsenyi2025influence} introduce a strongly conceptualised multi-agent system for replicating human reasoning in guessing games, where hypergames serve as a metaphor for players' nested beliefs and provide the mechanism for recursive reasoning. We extend this work by deriving a formal, generalised framework grounded in logic and propose a novel approach for recovering hypergame structures that rationalise unexpected outcomes. Our key contributions are as follows:
\begin{itemize}
    \item We formalise a multi-agent centralised hypergames framework that captures bounded rationality and enables theory-of-mind-like recursive reasoning.
    \item We express the system via a novel domain-specific language, formalising hypergame theory in a unified language grounded in logic.
    \item We introduce hypergame rationalisability via strong and weak equilibria as novel concepts, contributing to the expressiveness and analytical capabilities of the proposed language and framework.
    \item We revise the role of the umpire as a pseudo-player, rationalising target strategy profiles, materialised as an answer-set programming pipeline capable of recovering belief structures represented as hypergames that justify unexpected outcomes.
    \item We present two case studies that illustrate how the proposed language captures bounded rationality and nuanced strategic reasoning, while our pipeline maintains computational tractability.
\end{itemize}
In the following sections of the paper, we will first define our multi-agent system framework using subjective games and introduce the corresponding rationalizability criteria in Section~\ref{sec:mach}. Then, in Section~\ref{sec:DSL}, we will propose a minimal vocabulary for capturing hypergames and outlining the rationalisation process. We will also present logic programs designed to solve illustrative strategic scenarios where modelling misaligned perceptions is critical. Finally, we will discuss related works, summarise our contributions, and highlight future research objectives that emerge from this study.

\section{Hypergame Rationalisability}\label{sec:mach}
Hierarchical hypergames~\cite{Wang1988} provide an extended formalism and integrate higher-order expectations with the composite games of hypergame theory. Players may form expectations about the perceived game's components, where the order of the expectation marks the depth of reasoning, such as: \textit{Alice thinks that Bob believes that Cecil has...}. Accordingly, hypergames are specified by the highest order of expectation involved. Level 0 contains the ground-truth game \(G^*\) that an outside observer could specify if every player had complete information and identical preferences. Level 1 then records each player’s private view of that situation -- its \textit{subjective game} -- capturing misperceptions about actions, payoffs, or even the roster of participants. Level 2, in turn, models what every player believes about every \textit{other} player’s subjective game, and so on. Formally, a level-\(L\) hypergame is a tuple whose elements
are themselves hypergames of level at most \(L-1\); the construction bottoms out at \(L=0\) with ordinary \textit{base games}.

We propose a revised version of the centralised multi-agent interaction protocol from~\cite{trencsenyi2025approximating,trencsenyi2025influence}, inspired by the authority-mediated prisoner's dilemma~\cite{Burns1974structureofpd} and tailored to normal form games. The centralised, authoritative setting serves multiple purposes:
\begin{itemize}
    \item Isolating players enforces the non-cooperative nature of the interaction~\cite{NashJohn1951NG}.
    \item Implementing a central agent allows for various layers of functionality and representations: from a game facilitating mechanism, through a Nature-like model of structured uncertainty~\cite{Rasmusen1990-RASGAI}, to an explicit model of external influence~\cite{Burns1974structureofpd}.
    \item It realises a feedback loop that may accommodate validation, correction, or reinforcement mechanisms.
\end{itemize}
In the following subsections, we provide a formalisation for exploring hypergame structures that represent how players may develop misaligned interpretations of a base strategic scenario, through the lens of the umpire, a process we denote as \textit{Hypergame Rationalisation}.

\subsection{Multi-Agent Centralised Hypergames}

The framework by~\cite{trencsenyi2025approximating} introduces the umpire as an intermediary between isolated players. It is a non-strategic participant responsible for orchestrating the interaction and validating player moves. In this work, we expand this notion and define the umpire $\mathcal{U}$ as a pseudo-player whose goal is to rationalise a given strategy profile $a^*$ -- a tuple of players' actions -- with respect to a base game $G^*$. Formally, let 
\begin{equation}\label{eq:umpire}
    \mathcal{U}:(G^*, a^*) \mapsto \mathcal{H},
\end{equation}
where $\mathcal{H}$ is a set of hypergame structures where $a^*$ satisfies the hypergame rationalizability criteria. 

In order for the umpire $\mathcal{U}$ to rationalise a given strategy profile within a hypergame structure, it must operate relative to a well-defined strategic environment. This environment is specified by the \textit{base game} $G^*$, which provides the formal rules of interaction: the set of players, the available options, and the mapping from joint actions to payoffs.

\subsubsection{Base Game}\label{sec:basegame}
The base game constitutes the reference point from which subjective interpretations, and ultimately hypergame extensions, are constructed. We define base games as $G^*=(N,O,\Pi)$, where:
\begin{itemize}
    \item $N$ is the set of players;
    \item $O$ denotes the ``Options'', the set of all actions that exist in the context of G;
    \item $\Pi$ denotes the ``Outcomes'', it assigns each strategy profile $(a_i,\ldots, a_n) \in O \times O$ the corresponding payoffs $(\pi_i,\ldots,\pi_n)$ for $i \in N$, $\Pi: O^N \to \mathbb{R}^N$.
\end{itemize}
While in this work our objective is to evaluate hypothetical or post-hoc outcomes -- a process, in which \textit{players} are not active contributors -- for conceptual and contextual completeness we define them as $\mathcal{AG}=(\succ,I)$. A player is an agent characterised by an internal preference model $\succ$, which encapsulates individual traits -- such as social attitudes or preferences --, historical context, and, in general, all factors that influence the player's decision-making. Players form their interpretation of the base game via an interpretation function $I:(G^*,\succ) \to G^i$, spawning a subjective game.

\subsubsection{Subjective Game}\label{sec:subgame}

The subjective game $G^i=(N,A^i,\Pi^i)$\footnote{We assume interactions with no hidden actors, rendering $N$ as common knowledge.} is player $i$'s version of $G^*$, where:
\begin{itemize}
    \item \textbf{Actions:} $A^i$ is the subset of options player $i \in N$ believes each player $j \in N$ considers for this game: $A_j^i, \ldots, A_n^i$, where each $A^i_j \subseteq O$;\footnote{We assume $A^i \subseteq O$, bounding agents' beliefs to moves that exist in the context of $G^*$, and allowing us to omit invalid moves.}
    \item \textbf{Preferences:} $\Pi^i=(\succ_j^i,\ldots, \succ_n^i)$ is the joint preference rank\footnote{While $\Pi$ represents the payoff structure of $G^*$, $\Pi^i$ represents $i$'s subjective interpretation of payoffs in $G^i$ as ordinal ranks.}, where $\succ^i_j$ denotes what ranking $i$ thinks $j$ assigns to each strategy profile $(a_i,\ldots, a_n) \in A^i_j \times \ldots \times A^i_n$, for $i,j \in N$ -- with higher numbers denoting more preferred outcomes, $\Pi^i:A^i_j \times \ldots \times A^i_n \to \mathbb{N}^N$.
\end{itemize}

\subsubsection{Hierarchy of Subjective Games}
The proposed definition supports hypergames up to level 2:
\begin{itemize}
    \item In a 0-level hypergame, we assume complete information. We can collapse all $G^i$, taking $G^*$ as common knowledge;
    \item In a first-level hypergame $H^1$, misperceptions emerge, but players are not aware of them. Thus, player $i \in N$ expects the same rationality from others: $A^i_i = A^i_j$ and $\succ^i_i = \succ^i_j$ for $j\neq i, \quad j \in N$;
    \item In a second-level hypergame $H^2$, misperceptions emerge, and players generate beliefs about each opponent's interpretations. As such, player $i \in N$ assigns beliefs specific to each $j \in N$ over $A^i=A_j^i \times \cdots \times A_n^i$ and $\Pi^i=(\succ_j^i,\ldots, \succ_n^i)$.
    \item Given above definitions, we have $H^0=G^*$, $H^1=\{G^i\}_{i \in N}$ and $H^2=\{G^i\}_{i \in N}$, with $H^1 \not\cong H^2$ due to distinct notions of the underlying $A^i$ and $\Pi^i$.
\end{itemize}

\subsubsection{Hypergame Solutions}
Let $a^* = (a^*_i, a^*_{-i})$ denote a strategy profile, where $a^*_i$ is the strategy chosen by player $i \in N$ and $a^*_{-i}$ denotes the strategy profile of all other players. We then extend a typical game-theoretic solution concept, the Nash Equilibrium (NE), to capture subjective games as follows.

\paragraph{Subjective Best Response}

Given a strategy profile $a^*$, $a^*_i$ is a subjective best response, if:

\begin{equation}\label{eq:bestresponse}
    \Pi^i(a^*_i,a^*_{-i}) \geq  \Pi^i(a_i,a^*_{-i}), \forall a_i \in A^i.
\end{equation}

\paragraph{Strong Hypergame Nash Equilibrium}

A strategy profile $a^*$ is a strong hyper-Nash equilibrium (s-HNE) iff it is an NE in each corresponding subjective game:
\begin{equation}\label{eq:shne}
    \forall i\in N, a^* \in \text{NE}(G^i),
\end{equation}
where $\text{NE}(G^i)$ denotes the set of outcomes that satisfy the NE criteria.

\paragraph{Weak Hypergame Nash Equilibrium}

The strategy profile $a^*$ is a weak hypergame Nash equilibrium (w-HNE), if each player's choice $a^*_i \in a^*$ belongs to a Nash Equilibrium in the corresponding subjective game $G^i$:

\begin{equation}\label{eq:whne}
    \forall i \in N, \exists a^*_i \in \text{NE}(G^i).
\end{equation}

\subsection{Domain Specific Language}\label{sec:DSL}

As developing a direct representation for hypergames through nested, composite games is both conceptually complex and computationally intensive, a dedicated domain-specific language (DSL) is not only a conceptual nicety but serves an essential purpose. Without a unified vocabulary, each subjective perspective needs to be handcrafted, which becomes impractical as the number of players, options, and nested beliefs increases. By introducing a minimal, declarative syntax grounded in logic, we achieve the following benefits: (i) computational efficiency, since the solver can automatically generate and constrain subjective subgames instead of exhaustively enumerating them; (ii) formal clarity, by providing a unified semantics that eliminates ambiguity in defining hypergame structures and equilibria; and (iii) extensibility, as the same vocabulary can model a wide range of strategic scenarios -- from standard 2x2 dilemmas to multi-level reasoning with bounded rationality and information asymmetry. The DSL serves not only as a representational tool but also as a mechanism to connect theoretical hypergame models with practical multi-agent simulations. 

\begin{table}[h]
\centering
\caption{Vocabulary for hypergame rationalisation of 2-player normal form games via subjective action space and preferences.}
\begin{tabular}{ll}
\toprule
\texttt{player(}\textit{i}\texttt{)} & agent \textit{i} is a player\\
\texttt{role(}\textit{i}\texttt{,}\textit{r}\texttt{)} & \textit{i} has role \textit{r} \\
\texttt{option(}\textit{r}\texttt{,}\textit{a}\texttt{)} & \textit{r} has move \textit{a} \\
\texttt{payoff(}\textit{a}\texttt{,}\textit{b}\texttt{,}\textit{u}\texttt{,}\textit{v}\texttt{)} &  row(col) move \textit{a}(\textit{b}) yields \textit{u}(\textit{v})\\
\midrule
\texttt{chosen(}\textit{a}\texttt{,}\textit{b}\texttt{)} & row chose \textit{a} and col chose \textit{b} \\
\texttt{nash(}\textit{a}\texttt{,}\textit{b}\texttt{)} & \textit{a} and \textit{b} constitutes a NE\\
\texttt{utility(}\textit{r}\texttt{,}\textit{a}\texttt{,}\textit{b}\texttt{,}\textit{u}\texttt{,}\textit{v}) & a \texttt{payoff} interpreted by \textit{r}\\
\texttt{action(}\textit{r}\texttt{,}\textit{a}\texttt{)} & \textit{r} believes to have \texttt{option} \textit{a}\\
\bottomrule
\end{tabular}
\label{tab:vocab}
\end{table}

We specify a minimal set of reserved keywords for defining 2-player normal form games in this context, as outlined in the first section of Table~\ref{tab:vocab}. We specify every agent \textit{i} that participates in the game as a \texttt{player}. Corresponding to the normal form representation, each player is associated with a \texttt{role}: \textit{row} or \textit{column}. Following hypergame terminology, a move available in the game for the given role is referred to as an \texttt{option}. Pairs of row and column options constitute an outcome of the game, captured by a \texttt{payoff} linking roles' moves to consequent rewards. Given the game definition $G^* = (N,O,\Pi)$:

\begin{itemize}
    \item $N = \{\text{Alice}, \text{Bob}\}$;
    \item $O = \{C, D\}$, where $C$ denotes 
    \emph{Cooperate} and $D$ denotes \emph{Defect};
    \item $\Pi : O^N \to \mathbb{R}^N$ is a function defined as
    \[
        \Pi(a_{\text{Alice}}, a_{\text{Bob}}) =
        \begin{cases}
            (3,3), & \text{if } (a_{\text{Alice}}, a_{\text{Bob}}) = (C,C), \\
            (0,5), & \text{if } (a_{\text{Alice}}, a_{\text{Bob}}) = (C,D), \\
            (5,0), & \text{if } (a_{\text{Alice}}, a_{\text{Bob}}) = (D,C), \\
            (1,1), & \text{if } (a_{\text{Alice}}, a_{\text{Bob}}) = (D,D);
        \end{cases}
    \]
\end{itemize}
we can specify the prisoner's dilemma via the following atoms:
\begin{verbatim}
    player(alice).
    player(bob).
    role(alice, row).
    role(bob, column).
    option(row, cooperate).
    option(row, defect).
    option(column, cooperate).
    option(column, defect).
    payoff(cooperate, cooperate, 3, 3).
    payoff(cooperate, defect, 0, 5).
    payoff(defect, cooperate, 5, 0).
    payoff(defect, defect, 1, 1).
\end{verbatim}
To maintain the functionality of the solver and the expressiveness of our vocabulary, we make the following set of modelling simplifications. As players, options, and payoffs uniquely describe a game for our DSL, the introduction of an encapsulating predicate \texttt{game} is not necessary. In addition, we specify the scope of the proposed DSL to the process of \textit{hypergame rationalisation}, where we can assume a stateless domain. Given the definition of $\mathcal{U}$ (Eq.~\ref{eq:umpire}), the scope is restricted to the immediate context of the game, and we assume that the options of the game define the widest exclusive set of moves players can consider. Thus, no further checks and constraints are required on the validity of player moves in the corresponding game -- e.g., adopting \texttt{legal} from GDL~\cite{genesereth2005general} would yield \texttt{option(}\textit{r}\texttt{,}\textit{a}\texttt{)}$\Leftrightarrow$\texttt{legal(}\textit{r}\texttt{,}\textit{a}\texttt{)} in this context.

Next, we address the specification of rationalisation constraints -- as shown in the second section of Table~\ref{tab:vocab}. The atom \texttt{chosen(}\textit{a}\texttt{,}\textit{b}\texttt{)} encodes a strategy profile, highlighting a pair of \textit{row} and \textit{column} actions at evaluation. We define NE as mutual best responses following Equation~\ref{eq:bestresponse} and~\ref{eq:shne}:
\begin{verbatim}
    row_best(A,B) :-
        utility(A,B,R,_),
        R = #max { R1 : utility(A1,B,R1,_) }.
    col_best(A,B) :- 
        utility(A,B,_,C), 
        C = #max { C1 : utility(A,B1,_,C1) }.
    nash(A,B) :- 
        row_best(A,B),
        col_best(A,B).
\end{verbatim}
Abusing game-theoretic terminology, we distinguish between \textit{payoffs} provided in the \textit{base game} and \textit{utilities} that represent the interpretation of payoffs in a player's \textit{subjective game} as ordinal preference rankings -- encoded in role-specific \texttt{utility} atoms. Similarly, following the definitions from Section~\ref{sec:basegame}, an \texttt{action} is a move \textit{r} believes to be available. The predicates we defined in Table~\ref{tab:vocab} and the equilibrium constraints defined above provide us with the necessary representations to investigate hypergame rationalisability in various games and scenarios. 

\begin{figure}[h]
    \centering
    \includegraphics[width=1\linewidth]{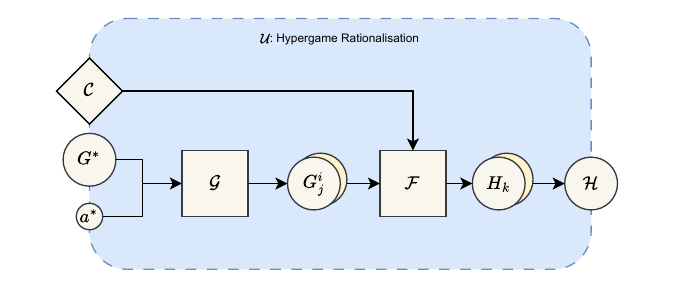}
    \caption{Hypergame rationalisation process for a game $G^*$ and players' selected strategies represented by a strategy profile $a^*$. Given $G^*$ and $a^*$, the umpire $\mathcal{U}$ generates ($\mathcal{G}$) all feasible subjective games. A filtering operation ($\mathcal{F}$) retains only those candidates where $a^*$ satisfies \textit{w}/\textit{s}-HNE -- or other constraints specified in $\mathcal{C}$. This operation then constructs the hypergame structures that comprise the resulting set $\mathcal{H}$.}
    \label{fig:rationalizer}
\end{figure}

Based on the game-independent definitions discussed so far, answer sets can be generated using additional game-dependent specifications for the considered action and utility spaces. This enables us to formalise the representation of agents' misaligned beliefs and provides a functional language that can generate such belief structures -- modelled as hypergames -- and validate the rationality of player moves. In the following subsections, we demonstrate how such game and context-dependent constraints can be implemented and evaluate the expressiveness and functionality of our pipeline via two illustrative examples: (1) modelling social attitudes and the emergence of mutual cooperation in the prisoner's dilemma, and (2) recovering belief structures compatible with the observed outcome in a complex real-world conflict situation. In Figure~\ref{fig:rationalizer}, we illustrate rationalising an arbitrary strategy profile $a^*$ for the corresponding base game $G^*$, which procedure is composed of the following elements:
\begin{description}
    \item[Generation:] Given a base game $G^*$ that defines all available moves as options $O$ and the corresponding payoff structure $\Pi$, and a strategy profile $a^*=(a_{\text{row}},a_{\text{column}})$, the generative operator $\mathcal{G}$ enumerates all possible subjective games $G^{\text{row}}_j$ that contains $a_\text{row}$ and $G^{\text{column}}_j$ that contains $a_\text{column}$.
    \item[Filtering:] Given a set of subjective games provided by $\mathcal{G}$ and a set of constraints $\mathcal{C}$, the filtering operator $\mathcal{F}$ composes a set of all hypergame structures that satisfy $\mathcal{C}$. $\mathcal{C}$ must entail a specification of a solution concept used for determining rationalisability -- such as w-HNE or s-HNE --, and optionally entails constraints on the considered action and utility space.
\end{description}

\section{Alignment via Bounded Rationality}
While the core assumptions of game theory do not account for the cognitive constraints, various extensions have been developed to address different sources of uncertainty in strategic interaction. In his theory of \textit{bounded rationality}, \cite{Simon1972} maintains the expectation of utility-maximizing rationality; however, he acknowledges that players are bounded by the information and time available for making decisions, as well as their ability to process information. Bounded rationality concepts capture agent heterogeneity, in which individual preferences and the social context are among the main contributing factors. 

\subsection{Cooperation as a Rational Solution in the Prisoner's Dilemma}

In their structural theory of social attitudes, ~\cite{burns1973structural} states that interpersonal reciprocity depends on the social orientations between actors, presenting four core attitudes that determine the structure of the interaction:
\begin{itemize}
    \item \textbf{Self-orientation}: an individualistic actor only cares about satisfying its personal goals with complete disregard or neutrality for any other actor's gains or losses.
    \item \textbf{Positive other-orientation}: an altruistic actor chooses actions that benefit another actor's goals. The social context of this orientation is that of belonging to a group whose goals coincide with the actor's individual values. Its individual goals are secondary in priority.
    \item \textbf{Negative other-orientation}: a hostile actor prefers to act in such a way that produces the most dissatisfaction for another actor.
    \item \textbf{Joint positive self/other-orientation}: a cooperative actor evaluates available actions and outcomes based on both its own goals and the goals of another actor.
\end{itemize}
One of the primary motivations behind Burns' work is the assertion that game-theoretic abstractions often overlook the social context of interactions, which can lead to the dehumanisation of the models. In light of these concerns, the proposed structural theory is also assessed within the framework of the prisoner's dilemma~\cite{Burns1974structureofpd}. This model is used to demonstrate how varying attitudes can be formed and how expected outcomes can change as a result. Accordingly, let $G^*_{\text{PD}}=(N,O,\Pi)$ denote a typical game of prisoner's dilemma, where:
\begin{itemize}
    \item $N=\{1,2\}$ is the set of two players;
    \item $O=\{C,D\}$ is the set of options, consisting of two actions: Cooperate (C) and Defect (D);
    \item $(a_i,a_{-i})=\begin{cases}
        T & \text{if } a_i=D,a_{-i}=C\\
        R & \text{if } a_i=C,a_{-i}=C\\
        P & \text{if } a_i=D,a_{-i}=D\\
        S & \text{if } a_i=C,a_{-i}=D\\
    \end{cases}$;
    \item $\Pi=\{T:(5,0),R:(3,3),P:(1,1),S:(0,5)\}$.
\end{itemize}
Given standard game-theoretic rational players, we have a shared preference relation over ordinal outcomes $\mathbb{O}=\{T > R > P >S \}$. $\mathbb{O}$ in this case is aligned with the specification of players with mutual \textit{self-oriented} attitudes, expected to terminate the game with the NE solution: $(D,D)$.

While altruism~\cite{axelrod} and byproduct mutualism~\cite{WORDEN2007411} are typical concepts that seek to explain the emergence of cooperation in the prisoner's dilemma, we propose the use of hypergames that integrate social attitudes to achieve the same effect. As players characterised by different social attitudes will interpret cardinal payoffs differently, their corresponding ordinal rankings of the dilemma's outcomes may diverge from the standard expectations. As such, we explore the hypergame rationalisation of mutual cooperation through the enumeration of subjective games where $(C,C)$ constitute an s-HNE (Eq.~\ref{eq:shne}). We assume common knowledge of the prisoner's dilemma game's options and thus limit the search to subjective utility structures. In order to restrict the solution space and ground preferences in the proposed structure of social attitudes, we constrain generated outcome rankings to be consistent with at least one of the corresponding preferences:
\begin{equation}
    \mathbb{O}_{\text{SO}}=\{T > R > P >S \},
\end{equation}
\begin{equation}
    \mathbb{O}_{\text{PO}}=\{S > R > P > T \},
\end{equation}
\begin{equation}
    \mathbb{O}_{\text{NO}}=\{T > P > R > S \},
\end{equation}
\begin{equation}
    \mathbb{O}_{\text{JO}}=\{R > T > P > S \},
\end{equation}
where $\mathbb{O}_{\text{SO}}$, $\mathbb{O}_{\text{PO}}$, $\mathbb{O}_{\text{NO}}$ and $\mathbb{O}_{\text{JO}}$ correspond to self-orientation, positive other-orientation, negative other-orientation and joint positive self/other-orientation in respective order. Note that here we implement $\mathbb{O}_\text{JO}$ following the multi-criteria decision-making model outlined in \cite{song1999fuzzy}, allowing us to omit cases of preferential indifference, with the assumption that a player with a joint positive-other orientation will prefer outcomes that maximise social welfare, while still prioritising their own individual utility maximisation. Specifying the attitude-specific constraints and the target strategy profile $a^*=(C,C)$, running our hypergame rationalisability solver yields four utility structures where mutual cooperation is an NE solution. The corresponding subjective games are presented in normal form in Figure~\ref{fig:four_matrices}.

\begin{figure}[ht]
    \centering
    \caption{Payoff matrices rationalising $(C,C)$ for base game $G^*_\text{PD}$ under different social attitude profiles. The payoff tuples are expressed as $(u_\text{row},u_\text{col})$, where $u_\text{row}$ is the row player's own ordinal preference ranking of the corresponding outcome and $u_\text{col}$ represents what the row player believes the opponent's preference ranking to be.}
    \begin{subfigure}{0.45\linewidth}
        \centering
        \caption{Expecting mutual positive other-orientation.}
        \begin{tabular}{c c c}
            \toprule
            & C & D \\
            \midrule
            C & $(3,3)$ & $(4,1)$ \\
            D & $(1,4)$ & $(2,2)$ \\
            \bottomrule
        \end{tabular}
        \label{fig:four_1}
    \end{subfigure}
    \hfill
    \begin{subfigure}{0.45\linewidth}
        \centering
        \caption{Joint positive self/other-orientation expecting positive other-orientation.}
        \begin{tabular}{c c c}
            \toprule
            & C & D \\
            \midrule
            C & $(4,3)$ & $(1,1)$ \\
            D & $(3,4)$ & $(2,2)$ \\
            \bottomrule
        \end{tabular}
        \label{fig:four_2}
    \end{subfigure}
    
    \vspace{1em}
    
    \begin{subfigure}{0.45\linewidth}
        \centering
        \caption{Expecting mutual joint positive self/other-orientation.}
        \begin{tabular}{c c c}
            \toprule
            & C & D \\
            \midrule
            C & $(4,4)$ & $(1,3)$ \\
            D & $(3,1)$ & $(2,2)$ \\
            \bottomrule
        \end{tabular}
        \label{fig:four_3}
    \end{subfigure}
    \hfill
    \begin{subfigure}{0.45\linewidth}
        \centering
        \caption{Positive other orientation expecting joint positive self/other-orientation.}
        \begin{tabular}{c c c}
            \toprule
            & C & D \\
            \midrule
            C & $(3,4)$ & $(4,3)$ \\
            D & $(1,1)$ & $(2,2)$ \\
            \bottomrule
        \end{tabular}
        \label{fig:four_4}
    \end{subfigure}
    \label{fig:four_matrices}
\end{figure}

Let $\mathcal{G}$ denote the set of candidate subjective games that rationalise $a^*$. $\mathcal{H}$ then enumerates hypergame structures that are composed of pairs of subjective games $G^i \in \mathcal{G}$. Having obtained the subjective games in which $(C,C)$ is an NE, we set $\mathcal{G}$ to contain the four matrices from Figure~\ref{fig:four_matrices}. We can now generate all hypergame hierarchies that rationalise mutual cooperation under s-HNE~( Eq.~\ref{eq:shne}).

\paragraph{Level-2 hypergames.}  
For each ordered pair of players $i,j\in N$ we construct a level-2 structure  
\begin{equation}
    H^2=\{G^i,G^j\}, \quad i,j\in N, \quad G^i,G^j \in \mathcal{G},
\end{equation}
Because $|\mathcal{G}|=4$ and $|N|=2$, this yields $4^{2}=16$ distinct level-2 hypergames: the $16$ permutations of the four subjective games shown in Figures~\ref{fig:four_1}–\ref{fig:four_4}. Each sustains $(C,C)$ as an s-HNE.

\paragraph{Level-1 hypergames.}  
In addition, if we adopt the standard assumption that rational players expect their opponents to reason as they do, we restrict attention to the two symmetric subjective games that embody either \textit{mutual positive–other orientation} or \textit{mutual joint self/other orientation}. We denote this subset by $\mathcal{G}'\subset\mathcal{G}$. The corresponding formation entails just $2^{2}=4$ level-1 hypergames: 
\begin{equation}
    H^1=\{G^i,G^j\}, \quad i,j\in N, \quad G^i,G^j \in \mathcal{G}'.
\end{equation}

\smallskip
\noindent In both constructions, every hypergame in $\mathcal{H}$ leads the players to prefer $(C,C)$ over the canonical defect–defect equilibrium of the base prisoner’s dilemma.

\subsection{Theory of Mind Explains the Fall of France}
While subjective games address the limitations set by assuming a common game with all players sharing the same perception of available strategies and payoffs, there may be more complex scenarios where an equilibrium concept can only be established once we consider players' reasoning about the opponents' model of the situation. Bennett~\cite{bennett1979complex} used a post-hoc study of the Fall of France scenario to illustrate the limitations of the ``common game'' assumption in a complex conflict and introduced \textit{cross-game information} as a notion for reasoning about other players' views. The Fall of France example is interesting from a hypergame analysis perspective because it entails information asymmetry and misaligned perceptions. The French High Command modelled the strategic situation as a two-option game: reinforcing the Maginot line or the North. While the French did not anticipate a German attack through the Ardennes due to the challenging terrain conditions, that is precisely what happened, leading to the catastrophic loss of the Allies, which was attributed to the High Command's strategic incompetence at the time.

\begin{figure}[ht]
    \centering
    \caption{French-German hypergame as a reconstruction of the Fall of France scenario by~\cite{bennett1979complex}. The action encoding is as follows: F1 -- Reinforce Maginot Line, F2 -- Reinforce the North, F3 -- Counter-attack Ardennes, G1 -- Attack Maginot Line, G2 -- Attack North, G3 -- Attack through Ardennes. Equilibrium outcomes are marked with bold text.}
    \begin{subfigure}{0.45\linewidth}
        \centering
        \caption{French Subjective Game}
        \label{fig:frenchsubgame}
        \begin{tabular}{c c c}
            \toprule
            & F1 & F2 \\
            \midrule
            G1 & $(1,4)$ & $(2,3)$ \\
            G2 & $(4,1)$ & $\mathbf{(3,2)}$ \\
            \bottomrule
        \end{tabular}
    \end{subfigure}
    \hfill
    \begin{subfigure}{0.45\linewidth}
        \centering
        \caption{German Subjective Game}
        \begin{tabular}{c c c c}
            \toprule
            & F1 & F2 & F3 \\
            \midrule
            G1 & $(1,4)$ & $(2,3)$ & $(2,3)$ \\
            G2 & $(4,1)$ & $(3,2)$ & $\mathbf{(3,2)}$ \\
            G3 & $(3,2)$ & $(5,0)$ & $(2,3)$ \\
            \bottomrule
        \end{tabular}
    \end{subfigure}
    \label{fig:francevgermany}
\end{figure}
The hypergame reconstruction of the scenario is presented in Figure~\ref{fig:francevgermany}. In contrast to the previous ruling, the equilibrium analysis of the French subjective game reveals that sending reinforcements to the North is a rational choice, given the available information. However, while these subjective games model the players' bounded rationality and explain the French strategy, they still do not capture the Germans' reasoning and decision to push through the Ardennes. In this instance, we can utilise our proposed hypergame rationalizability criteria to try to recover structures that rationalise the strategy profile $a^*=(G3,F2)$. We adopt the fixed utility structure of~\cite{bennett1979complex}, which assumes that both the French and German high commands evaluated the possible outcomes with equal competence, and thus what is ``good'' for one is equally ``bad'' for the other. This consideration sets our scope to subjective action spaces, where the French game is fixed to the two-option game presented in Figure~\ref{fig:frenchsubgame}. Given that the French subjective game does not include \textit{G3}, we must rely on the w-HNE (Equation~\ref{eq:whne}) as our rationalizability criteria. Then, the NE membership constraint is a simple extension to the previously introduced DSL, as follows:
\begin{verbatim}
    in_nash_row(A) :- nash(A,_).
    in_nash_col(B) :- nash(_,B).
\end{verbatim}
The resulting hypergame -- where \textit{F2} is part of a NE in the French subjective game, and \textit{G3} is part of a NE in the German subjective game -- is composed of the French game model that we already rationalised without cross-game information, and the sole w-HNE solution for Germany that rationalises \textit{G3}: $H^2=\{G^{\text{France}},G^{\text{Germany}}\}$, where $A^{\text{France}}_{\text{Germany}}=\{\text{G1,G2}\}$ and $A^{\text{Germany}}_{\text{France}}=\{\text{F2}\}$. This result is aligned with Bennet and Dando's~\cite{bennett1979complex} proposed solution, which assumes the Germans had intel about the French strategy, leaving Germany with a reduced subgame and the simple task of finding the best response for Allied forces moving North (\textit{F2}): a surprise attack through the Ardennes (\textit{G3}).

While ~\cite{vane2006advances} defines hypergames as models of bounded rationality, given that hypergames at $L>1$ explicitly capture recursive reasoning -- players' beliefs of others' games or hypergames --, we take Vane's statement further and specify hypergames as models of TOM. The Fall of France scenario is a perfect illustration of this capability, where the sole structure that rationalises the observed outcome entails a reasoning closely aligned with the original concept of \textit{false beliefs} discussed in the context of TOM~\cite{WIMMER1983theoryofmindexperiment}: \textit{I know what you (don't) know}.

\section{Related Work}

There are various approaches for formalising agent-based game-theoretic interactions, such as GALA~\cite{koller1997representations} or GDL and its incomplete information and epistemic extensions~\cite{genesereth2005general,thielscher2011gdl,thielscher2017gdl}, that attempt to ground gameplay in logic and provide machine-compatible general semantics. More recently, GDL-inspired logic programs have been used as ``solvers'' in game-theoretic agent-based contexts for representing games, simulating play, and validating strategic reasoning~\cite{mensfelt2024autoformalization,mensfelt2024autoformalizing,mensfelt2024logic}. However, for hypergame theory in particular, there are only preliminary or partial applications. HYPANT~\cite{brumley2003hypant} is a software tool that entails an XML-based hypergame-specific grammar (HML) for representing games and a hypergame analysis algorithm that managed to reconstruct the manual analytical results of selected hypergame-theoretic papers. HYPANT is unable to create and manipulate hypergame structures; data must be handcrafted and provided using HML. Similarly, the HAT software~\cite{Gibson2013hypergame} was developed to replace the tedious process of manual hypergame analysis. Instead of a formal grammar, it uses the Hypergame Normal Form -- a decision-theoretic take on hypergames developed by~\cite{vane2000thesis} -- expressed in pure XML to represent games. While HAT provides an additional layer of flexibility over HYPANT by introducing utility functions and the element of randomness, it still cannot generate hypergame structures, and the representation is limited to a single viewpoint at a time. Neither of the existing approaches fully implements a practical formal language that can be used to represent, generate, and solve higher-order hypergames capturing complex game-theoretic scenarios with players of bounded rationality or that provides the representational foundations for recursive reasoners.

\section{Conclusions}

We introduced a multi-agent-based simulation framework for game-theoretic interaction based on centralised hypergames that captures misaligned interpretations of a core game. We grounded the representation in logic, introducing a novel DSL that formalises and unifies hypergame-theoretic analysis. Empirical evidence from a social-attitude variant of the Prisoner’s Dilemma and a reconstruction of the Fall of France demonstrates that the framework both captures nuanced strategic reasoning and remains computationally tractable in practice. These results position hypergame theory as a practical tool for studying misalignment and deception in modern multi-agent AI systems, complementing existing game-theoretic and simulation-based approaches.

We identify several opportunities for future work that arise from this study, summarised as follows:
\begin{itemize}
    \item While the multi-agent centralised hypergames concept outlined in Section~\ref{sec:mach} accommodates any number of players, our current DSL and associated solvers are limited to two-player normal form games. A natural next step would be to extend the syntax to handle the $n$-player case and conduct case studies involving conflicts with more than two players.
    \item In this work, we linked the umpire to the process of hypergame rationalisation, which currently assumes a stateless environment and treats players' decisions as static components in the process. However, with an extended syntax, our proposed language has the potential to represent players as dynamic, recursive reasoners. By drawing inspiration from descriptive languages like GDL, a revised syntax could equip us with the tools necessary for expressing hypergame-based decision-making, moving us closer to a general-purpose hypergame description language.
    \item Once dynamic play is incorporated, we can elevate the umpire from an \textit{ex-post} rationalizer to an \textit{in-situ} ``referee'' capable of validating each announced move against the logic-based rules of the current subjective game, as well as injecting sanctions or corrective feedback when illegal or irrational actions are detected.
    \item Additionally, we could leverage the formalism of our proposed DSL in conjunction with recent LLM-driven \textit{auto-formalisation} pipelines~\cite{mensfelt2024autoformalization,mensfelt2024autoformalizing} to dynamically translate natural language descriptions of actions or scenario updates into our DSL.  
\end{itemize}
To the best of our knowledge, no prior work offers a unified, logic-grounded representation for subjective games together with a principled procedure for rationalising observed outcomes via compatible bounded-rationality structures. By establishing this foundation -- and by outlining concrete extensions in scalability, dynamism, and human-in-the-loop auto-formalisation -- we hope to catalyse a new research agenda at the intersection of game theory, multi-agent simulation, and AI alignment. Realising the roadmap sketched above would yield an expressive language paired with an autonomous simulator that supports rigorous, formally guaranteed analysis of strategic interactions in increasingly complex and realistic settings.

\begin{credits}
\subsubsection{\ackname} This work was supported by a Leverhulme Trust International Professorship Grant (LIP-2022-001).\\
The author would like to express his gratidude to Kostas Stathis for his constructive comments and fruitful discussions about logical representations of games.

\end{credits}

\bibliographystyle{splncs04}
\bibliography{references}
\end{document}